\documentclass[runningheads]{llncs}
\usepackage{graphicx}
\usepackage{mathtools}
\usepackage{amsmath}
\usepackage{subcaption}
\usepackage{booktabs}
\usepackage{tikz}

\DeclarePairedDelimiter\abs{\lvert}{\rvert}%

\renewcommand{\vec}[1]{\mathbf{#1}}

\begin{document}
\title{Linear and Deformable Image Registration \\ with 3D Convolutional Neural Networks}
\titlerunning{Linear and Deformable Image Registration with 3D CNNs}
\author{Christodoulidis Stergios\inst{1}, Sahasrabudhe Mihir\inst{2}, Vakalopoulou Maria\inst{2}, Chassagnon Guillaume\inst{2,3}, Revel Marie-Pierre\inst{3}, Mougiakakou Stavroula\inst{1}, Paragios Nikos\inst{4}}
\authorrunning{S. Christodoulidis et al.}
\institute{ARTORG Center, University of Bern, Murtenstrasse 50, 3008 Bern, Switzerland \\
\email{\{stergios.christodoulidis, stavroula.mougiakakou\}@artorg.unibe.ch} \\
\and CVN, CentraleSup\'{e}lec, Universit\'{e} Paris-Saclay, 91190 Gif-sur-Yvette, France \\
\email{\{mihir.sahasrabudhe, maria.vakalopoulou, guillaume.chassagnon\}@centralesupelec.fr}
\and Groupe Hospitalier Cochin-H\^{o}tel Dieu, Universit\'{e} Paris Descartes, France \\
\email{marie-pierre.revel@aphp.fr}
\and TheraPanacea, Paris, France \\
\email{n.paragios@therapanacea.eu}}
\maketitle              % typeset the header of the contribution
\begin{abstract}
Image registration and in particular deformable registration methods are  pillars of medical imaging. Inspired by the recent advances in deep learning, we propose in this paper, a novel convolutional neural network architecture that couples linear and deformable registration within a unified architecture endowed with near real-time performance. Our framework is modular with respect to the global transformation component, as well as with respect to the similarity function while it guarantees smooth displacement fields. We evaluate the performance of our network on the challenging problem of MRI lung registration, and demonstrate superior performance with respect to state of the art elastic registration methods. The proposed deformation (between inspiration \& expiration) was considered within a clinically relevant task of interstitial lung disease (ILD) classification and showed promising results.

\end{abstract}
\section{Introduction}
Image registration is the process of aligning two or more sources of data to the same coordinate system. Through all the different registration methods used in medical applications, deformable registration is the one most commonly used due to its richness of description~\cite{Sotiras:13}. The goal of deformable registration is to calculate the optimal non-linear dense transformation $G$ to align in the best possible way, a source (moving) image $S$ to a reference (target) image $R$~\cite{Ferrante:16,avants:08}. Existing literature considers the mapping once the local alignment has been performed and therefore is often biased towards the linear component.  Furthermore, state of the art methods are sensitive to the application setting, involve multiple hyper-parameters (optimization strategy, smoothness term, deformation model, similarity metric) and are computationally expensive. 

Recently, deep learning methods have gained a lot of attention due to their state of the art performance on a variety of problems and applications~\cite{8237808,Guler2018DensePose}. In computer vision, optical flow estimation---a problem highly similar to deformable registration---has been successfully addressed with numerous deep neural network architectures~\cite{hui18liteflownet}. In medical imaging, some methods in literature propose the use of convolutional neural networks (CNNs) as robust methods for image registration~\cite{Simonovsky:15,Cheng:16}. More recently, adversarial losses have been introduced with impressive performance~\cite{Yan:18}. The majority of these methods share two limitations: \textit{(i)} dependency on the linear component of the transformation and \textit{(ii)} dependency on ground truth displacement which is used for supervised training.

In this paper, we address the previous limitations of traditional deformable registration methods and at the same time propose an unsupervised method for efficient and accurate registration of 3D medical volumes that determines the linear and deformable parts in a single forward pass. The proposed solution outperforms conventional multi-metric deformable registration methods and demonstrates evidence of clinical relevance that can be used for the classification of patients with ILD using the transformation between the extreme moments of the respiration circle.

The main contributions of the study are fourfold: \textit{(i)} coupling linear and deformable registration within a single optimization step / architecture, \textit{(ii)} creating a modular, parameter-free implementation which is independent of the different similarity metrics, \textit{(iii)} reducing considerably the computational time needed for registration allowing real-time applications, \textit{(iv)} associating deformations with clinical information.

\section{Methodology}

In this study, we propose the use of an unsupervised CNN for the registration of pairs of medical images. A source image $S$ and a reference image $R$ are presented as inputs to the CNN while the output is the deformation $G$ along with the registered source image $D$. This section presents details of the proposed architecture as well as the dataset that we utilized for our experiments. Please note that henceforth, we will use the terms \emph{deformation}, \emph{grid}, and \emph{transformation} interchangeably.

\subsection{Linear and Deformable 3D Transformer}

One of the main components of the proposed CNN is the 3D transformer layer. This layer is part of the CNN and is used to warp its input under a deformation $G$. The forward pass for this layer is given by

\begin{equation}
D = \mathcal{W}(S, G),
\end{equation}

where $\mathcal{W}(\cdot, G)$ indicates a sampling operation $\mathcal{W}$ under the deformation $G$. $G$ is a dense deformation which can be thought of as an image of the same size as $D$, and which is constructed by assigning for every output voxel in $D$, a sampling coordinate in the input $S$.

In order to allow gradients to flow backwards though this warping operation and facilitate back-propagation training, the gradients with respect to the input image as well as the deformation should be defined. Similar to~\cite{jaderberg2015spatial}, such gradients can be calculated for a backward trilinear interpolation sampling. The deformation is hence fed to the transformer layer as sampling coordinates for backward warping. The sampling process is illustrated by

\begin{equation}
D(\vec{p}) = \mathcal{W}(S, G)(\vec{p}) = \sum_{\vec{q}} S(\vec{q}) \prod_{d}\max\left(0, 1 - \abs{[G(\vec{p})]_d - \vec{q}_d}\right),
\label{eq:sampling}
\end{equation}

where $\vec{p}$ and $\vec{q}$ denote pixel locations, $d\in\{x,y,z\}$ denotes an axis, and $[G(\vec{p})]_d$ denotes the $d$-component of $G(\vec{p})$.

Our modeling of the deformation $G$ offers a choice of the type of deformation we wish to use---linear, deformable, or both. The linear (or affine) part of the deformation requires the prediction of a $3 \times 4$ affine transformation matrix $A$ according to the relation $[\hat{x}, \hat{y}, \hat{z}]^T = A[x, y, z, 1]^T $, where $[x, y, z, 1]^T$ represents the augmented points to be deformed, whereas $[\hat{x}, \hat{y}, \hat{z}]^T$ represents their locations in the deformed image. The matrix $A$ can then be used to build a grid, $G_A$, which is the affine component of the deformation $G$.

\iffalse
\begin{equation}
\begin{bmatrix}
\hat{x} \\ 
\hat{y} \\ 
\hat{z} 
\end{bmatrix}
= A\begin{bmatrix}
x \\ y  \\ z  \\ 1
\end{bmatrix} = 
\begin{bmatrix}
a_{00} & a_{01} & a_{02} & a_{03}\\ 
a_{10} & a_{11} & a_{12} & a_{13}\\ 
a_{20} & a_{21} & a_{22} & a_{23}
\end{bmatrix}
\begin{bmatrix}
x \\ y  \\ z  \\ 1
\end{bmatrix}, 
\end{equation}
\fi

To model the deformable part $G_N$, a simple and straightforward approach is to generate sampling coordinates for each output voxel ($G_N(\vec{p})$). We can let the network calculate these sampling points directly. Such a choice would however require the network to produce feature maps with large value ranges which complicates training. Moreover without appropriate regularization, non-smooth and even unconnected deformations could be produced. In order to circumvent this problem, we adopt the approach proposed by \cite{shu:18} and predict spatial gradients $\Phi$ of the deformation along each dimension instead of the deformation itself. This quantity measures the displacements of consecutive pixels. By enforcing these displacements to have positive values and subsequently applying an integration operation along each dimension, the spatial sampling coordinates can be retrieved. This integration operation could be approximated by simply applying a cumulative sum along each dimension of the input (i.e integral image). In such a case, for example, when $\Phi_{\vec{p}_d} = 1$ there is no change in the distance between the pixels $\vec{p}$ and $\vec{p}+1$ in the deformed image along the axis $d$. On the other hand, when $\Phi_{\vec{p}_d} < 1$, the distance between these consecutive pixels along $d$ will decrease, while it will increase when $\Phi_{\vec{p}_d} > 1$. Such an approach ensures the generation of smooth deformations that avoid self-crossings, while allows the control of maximum displacements among consecutive pixels. 

Finally, to compose the two parts we apply the deformable component to a moving image, followed by the linear component. When operating on a fixed image $S$, this step can be written as

\begin{equation}
	\mathcal{W}(S, G) = \mathcal{W}\left(\mathcal{W}(S, G_N), G_A\right).
\end{equation}

During training, the optimization of the decoders of $A$ and $G_N$ is done jointly, as the network is trained end-to-end. We also impose regularization constraints on both these components. 
We elaborate on the importance of this regularization for the joint training in Section~\ref{sec:training}.

\subsection{Architecture}

\begin{figure}[t!]
\centering
\includegraphics[width=1\textwidth]{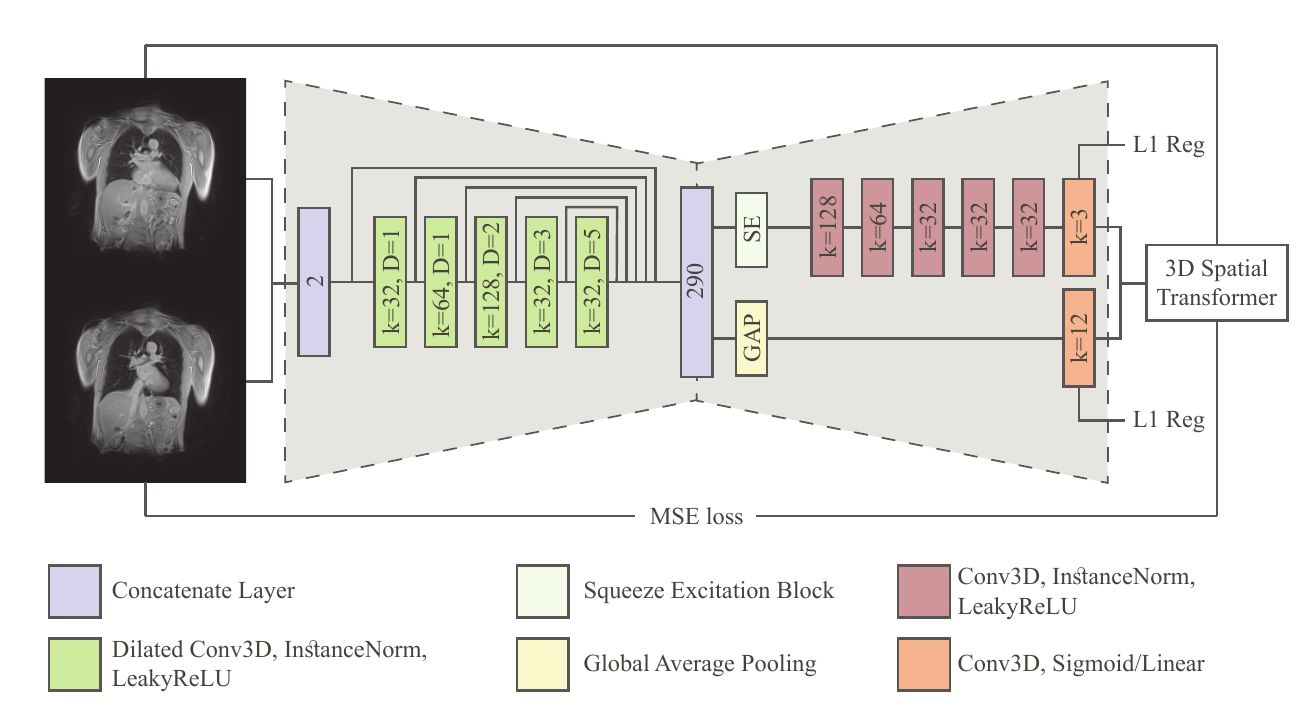}
\caption{The overall CNN architecture. The network uses a pair of 3D images and calculates the optimal deformations from the one image to the other.}
\label{fig:architecture}
\vspace{-10pt}
\end{figure}

The architecture of the CNN is based on an encoder-decoder framework presented in~\cite{anthimopoulos2018semantic} (Fig.~\ref{fig:architecture}). The encoder adopts dilated convolutional kernels along with multi-resolution feature merging, while the decoder employs non-dilated convolutional layers and up-sampling operations. Specifically, a kernel size of $3 \times 3 \times 3$ was set for the convolutional layers while LeakyReLU activation was employed for all convolutional layers except the last two. Instance normalization was included before most of the activation functions. In total five layers are used in the encoder and their outputs are merged along with the input pair of image to form a feature map of 290 features with a total receptive field of $25 \times 25 \times 25$. In the decoder, two branches were implemented---one for the spatial deformation gradients and the other for the affine matrix. As far as the former is concerned, a squeeze-excitation block~\cite{hu2017squeeze} was added in order to weigh the most important features for the spatial gradients calculation while for the latter a simple global average operation was used to reduce the spatial dimensions to one. For the affine parameters and the spatial deformation gradients, a linear layer and sigmoid activation were respectively used. Finally to retrieve $\Phi$, the output of the sigmoid function should be scaled by a factor of 2 in order to fall in the range $[0,2]$ and hence allow for consecutive pixels to have larger distance than the initial.

\subsection{Training}
\label{sec:training}
The network was trained by minimizing the mean squared error (MSE) between the $R$ and $D$ image intensities as well as the regularization terms of the affine transformation parameters and the spatial deformation gradients using the Adam optimizer~\cite{kingma2014adam}. Our loss is defined as

\begin{equation}
\mathrm{Loss} = \left \| R - \mathcal{W}(S, G) \right \|^{2} + \alpha \left\| A - A_I\right\|_{1} + \beta \left \| \Phi - \Phi_{I} \right \|_{1},
\end{equation}

where $A_I$ represents the identity affine transformation matrix, $\Phi_{I}$ is the spatial gradient of the identity deformation, and $\alpha$ and $\beta$ are regularization weights. As mentioned before, regularization is essential to the joint optimization. To elaborate, without the L1 regularization on $A$, the network might get stuck in a local minimum where it aligns only high-level features using the affine transformation. This will result in a high reconstruction error. On the other hand, without the smoothness regularizer on $\Phi$, the spatial gradients decoder network can predict very non-smooth grids which again makes it prone to fall in a local minimum. Having both linear and deformable components is helpful to the network because these two components now share the work. This hypothesis aligns with \cite{shu:18} and is also evaluated in Section \ref{sec:experiments}.

The initial learning rate is $10^{-3}$ and subdued by a factor of $10$ if the performance on the validation set does not improve for 50 epochs while the training procedure stops when there is no improvement for 100 epochs. The regularization weights $\alpha$ and $\beta$ were set to $10^{-6}$ so that neither of the two components has an unreasonably large contribution to the final loss. As training samples, random pairs among all cases were selected with a batch size of $2$ due to the limited memory resources on the GPU. The performance of the network was evaluated every $100$ batches, and both proposed models---with and without affine components---converged after nearly $300$ epochs. The overall training time was calculated to $\sim16$ hours.

\subsection{Dataset}
MRI exams were acquired as a part of a prospective study aiming to evaluate the feasibility of pulmonary fibrosis detection in systemic sclerosis patients by using magnetic resonance imaging (MRI) and an elastic registration-driven biomarker. This study received institutional review board approval and all patients gave their written consent. The study population consisted of 41 patients (29 patients with systemic sclerosis and 12 healthy volunteers). Experienced radiologists annotated the lung field for the total of the 82 images and provided information about the pathology of each patient (healthy or not). Additionally, eleven characteristic landmarks inside the lung area had been provided by two experienced radiologists.

All MRI examinations were acquired on a 3T-MRI unit (SKYRA magneton, Siemens Healthineers) using an 18-phased-array-body coil. All subjects were positioned in the supine position with their arms along the body. Inspiratory and expiratory MRI images were acquired using an ultrashort time of echo (UTE) sequence, the spiral VIBE sequence, with the same acquisition parameters (repetition time $2.73$ ms, echo time $0.05$ ms, flip angle $5^{\circ}$, field-of-view $620 \times 620$ mm, slice thickness 2.5 mm, matrix $188 \times 188$, with an in-plane resolution of $2.14 \times 2.14$ mm).

As a pre-processing step, the image intensity values were cropped within the window $[0, 1300]$ and mapped to $[0, 1]$. Moreover, all the images were scaled down along all dimensions by a factor of $2/3$ with cubic interpolation resulting to an image size of $64\times192\times192$ to compensate GPU memory constraints. A random split was performed and 28 patients (56 pairs of images) were selected for the training set, resulting to $3136$ training pairs, while the rest 13 were used for validation.

\section{Experimental Setup and Results}
\label{sec:experiments}
\subsection{Evaluation}
We evaluated the performance of our method against two different state-of-the-art methods, namely, Symmetric Normalization (SyN)~\cite{avants:08}, using its implementation on the ANTs package~\cite{avants:11} and the deformable method presented in~\cite{glocker:11,Ferrante:16} for a variety of similarity metrics (normalized cross correlation (NCC), mutual information (MI) and discrete wavelet metric (DWM), and their combination). For the evaluation we calculated the Dice coefficient metric, measured on the lung masks, after we applied the calculated deformation on the lung mask of the moving image. Moreover, we evaluate our method using the provided landmark locations. For comparison reasons we report the approximate computational time each of these methods needed to register a pair of images. For all the implementations we used a GeForce GTX 1080 GPU except for SyN implementation where we used a CPU implementation running on $4$ cores of an i7-4700HQ CPU.

\subsection{Results and Discussion}
Starting with the quantitative evaluation, in Table~\ref{tab2} the mean Dice coefficient values along with their standard deviations are presented for different methods. We performed two different types of tests. In the first set of experiments (Table~\ref{tab2}: Inhale-Exhale), we tested the performance of the different methods for the registration of the MRI images, between the inhale and exhale images, for the 13 validation patients. The SyN implementation reports the lowest Dice scores while at the same time, it is computationally quite expensive due to its CPU implementation. Moreover, we tested three different similarity metrics along with their combinations using the method proposed in~\cite{Ferrante:16} as described earlier. In this specific setup, the MI metric seam to report the best Dice scores. However, the scores reported by the proposed architecture are superior by at least $\sim2.5$\% to the ones reported by the other methods. For the proposed method, the addition of a linear component to the transformation layer does not change the performance of the network significantly in this experiment. Finally, we calculated the errors over all axes in predicted locations for eleven different manually annotated landmark points on inhale volumes after they were deformed using the decoded deformation for each patient. We compare the performance of our method against the inter-observer (two different medical experts) distance and the method presented in~\cite{Ferrante:16} in Table \ref{tab:landmark}. We observe that both methods perform very well considering the inter-observer variability, with the proposed one reporting slitly better average euclidean distances.

\begin{table}[t!]
	\centering
	\caption{Dice coefficient scores (\%) calculated over the deformed lung masks and the ground truth.}
    \vspace{4pt}
	\label{tab2}
	\begin{tabular*}{\linewidth}{l@{\extracolsep{\fill}} r r r}
		\toprule
        \multicolumn{1}{c}{Method} &
        \multicolumn{1}{c}{Inhale-Exhale} &
		\multicolumn{1}{c}{All Combinations} &
        \multicolumn{1}{c}{Time/subject (s)}  \\
		\midrule
		Unregistered & 75.62$\pm$10.89 & 57.22$\pm$12.90 & $-$ \\
        \addlinespace[7pt]
		Deformable with NCC~\cite{Ferrante:16}  & 84.25$\pm$6.89&76.10$\pm$7.92& $\sim$1 (GPU) \\
		Deformable with DWM~\cite{Ferrante:16} & 88.63$\pm$4.67& 75.92$\pm$8.81 & $\sim$2 (GPU) \\
		Deformable with MI~\cite{Ferrante:16} & 88.86$\pm$5.13 & 76.33$\pm$8.74 & $\sim$2 (GPU)\\
		Deformable with all above~\cite{Ferrante:16} & 88.81$\pm$5.85&78.71$\pm$8.56 & $\sim$2 (GPU)\\
        SyN~\cite{avants:08}  & 83.86$\pm$6.04& $-$ & $\sim$2500 (CPU)\\
        \addlinespace[7pt]
		Proposed w/o Affine & 91.28$\pm$2.47 & 81.75$\pm$7.88 & $\sim$0.5 (GPU)\\
		Proposed & \textbf{91.48}$\pm$\textbf{2.33} & \textbf{82.34}$\pm$\textbf{7.68} & $\sim$0.5 (GPU)\\
		\bottomrule
	\end{tabular*}
\end{table}

For the second set of experiments (Table~\ref{tab2}: All combinations), we report the Dice scores for all combinations of the 13 different patients, resulting on $169$ validation pairs. Due to the large number of combinations, this problem is more challenging since the size of the lungs in the extreme moments of the respiratory circles can vary significantly. Again, the performance of the proposed architecture is superior to the tested baselines, highlighting its very promising results. In this experimental setup, the linear component plays a more important part by boosting the performance by $\sim0.5$\%.

Concerning the computation time, both~\cite{Ferrante:16} and the proposed method report very low inference time, due to their GPU implementations, with the proposed method reaching  $\sim0.5$ seconds per subject. On the other hand,~\cite{avants:08} is computationally quite expensive, making it difficult to test it for all the possible combinations on the validation set.

\begin{table}[ht!]
\centering
\caption{Errors measured as average euclidean distances between estimated landmark locations and ground truth marked by two medical experts. We also report as \emph{inter-observer}, the average euclidean distance between same landmark locations marked by the two experts. $dx$, $dy$, and $dz$ denote distances along $x$-, $y$-, and $z$- axes, respectively, while $ds$ denotes the average error along all axes.}
\label{tab:landmark}
\begin{tabular}{p{7cm} p{1cm} p{1cm} p{1cm} p{1cm}}
\toprule
Method            & $dx$ & $dy$ & $dz$ & $ds$ \\
\midrule
Inter-observer    &   1.664  & 2.545   & 1.555   & 3.905   \\
\addlinespace[7pt]
Deformable with NCC, DWM, and MI~\cite{Ferrante:16} &  1.855  & 3.169   &  2.229  & 4.699   \\
\addlinespace[7pt]
Proposed w/o Affine         & 2.014   &  2.947  &  1.858  &  4.569  \\
Proposed &  \textbf{1.793}  & \textbf{2.904}   &  \textbf{1.822}  & \textbf{4.358}   \\

\bottomrule
\end{tabular}
\end{table}

\begin{figure}[b]
		\centering
        \begin{subfigure}[b]{0.23\textwidth}
                \centering
                \includegraphics[height=2.8cm]{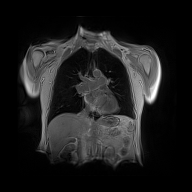}
                \caption{Reference image}
        \end{subfigure}%
        ~
        \begin{subfigure}[b]{0.23\textwidth}
        		\centering
				\includegraphics[height=2.8cm]{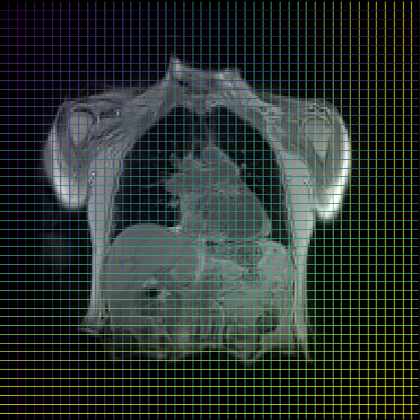}
                \caption{Moving image}
        \end{subfigure}%
        ~
        \begin{subfigure}[b]{0.23\textwidth}
        		\centering
                \includegraphics[height=2.8cm]{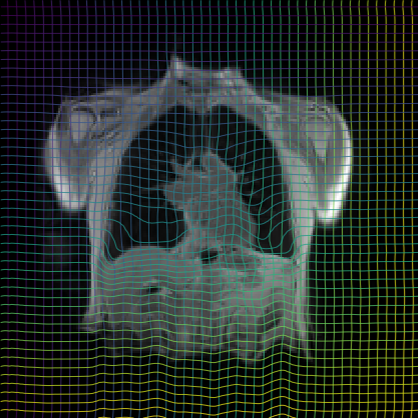}
                \caption{Deformed image}
        \end{subfigure}%
        ~
        \begin{subfigure}[b]{0.3\textwidth}
        		\centering
                \includegraphics[height=2.8cm]{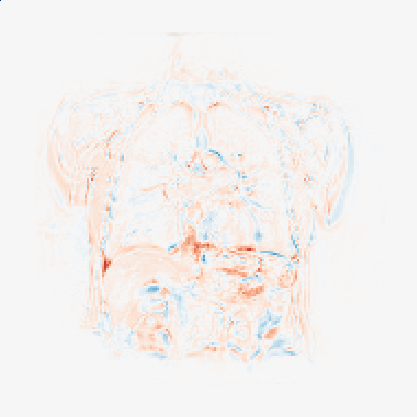}
                \includegraphics[height=2.8cm]{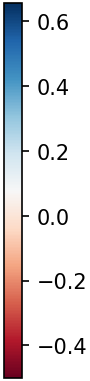}
                \caption{Difference}
        \end{subfigure} 
        \caption{A visualized registration of a pair of images, generated by the proposed architecture. The initial and deformed grids are superimposed on the images.}\label{fig:examples1}
        \vspace*{-2mm}
\end{figure}

Finally, in Figure~\ref{fig:examples1}, we present the deformed image produced by the proposed method on coronal view for a single patient in the two different moments of the respiratory cyrcle. The grids were superimposed on the images, indicating the displacements calculated by the network. The last column shows the difference between the reference and deformed image. One can observe that the majority of the errors occur on the boundaries, as the network fails to capture large local displacements.

\subsection{Evaluation of the Clinical Relevance of the Deformation}

To asses the relevance of the decoded transformations in a clinical setting, we trained a small classifier on top of the obtained residual deformations to classify patients as healthy or unhealthy. The residual deformation associated with a pair of images indicates voxel displacements, written as $G_\delta = G - G_I$, where $G$ is the deduced deformation between the two images, and $G_I$ is the identity deformation.

\begin{figure}[ht!]
	\centering
    \includegraphics[width=0.7\linewidth]{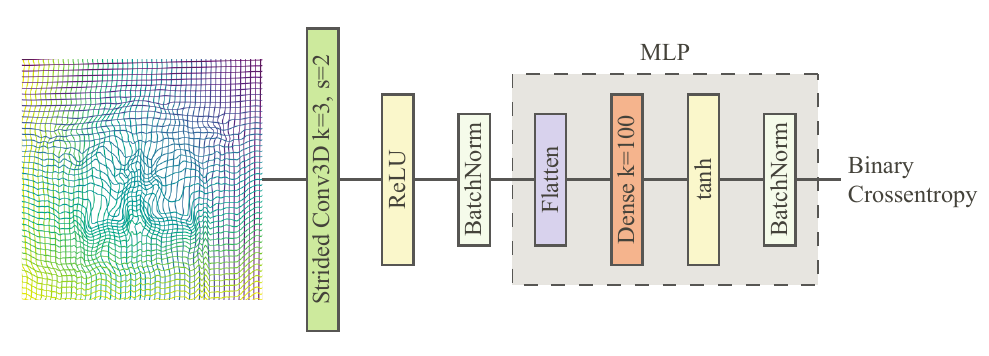}
    \caption{The neural network trained as a classifier on top of the transformations.}%
	\label{fig:NNclassifier}
\end{figure}

We trained a downsampling convolutional kernel followed by a multi-layer perceptron (MLP) to be able to predict whether a case is healthy or not. The network architecture is shown in Figure \ref{fig:NNclassifier}. The model includes batch normalization layers, to avoid overfitting, as we have few training examples at our disposal. Further, a Tanh activation function is used in the MLP. The downsampling kernel is of size $3 \times 3 \times 3$, with a stride of $2$ and a padding of $1$. The number of units in the hidden layer of the MLP was set to $100$. We trained with binary cross entropy loss, with an initial learning rate of $10^{-4}$, which is halved every fifty epochs. Training five models in parallel took about $2$ hours on two GeForce GTX 1080 GPUs.

We cross-validate five models on the training set of 28 patients, and report the average response of these models on the rest 13 patients. We conduct the same experiment for deformations obtained using~\cite{Ferrante:16} and all similarity measures (NCC, DWM, MI). The results on the test set using a threshold of $0.5$ on the predicted probability are reported in Table~\ref{tab:disease}, suggesting that indeed the deformations between inhale and exhale carry information about lung diseases.

\begin{table}[ht!]
	\centering
    \caption{Results on disease prediction using deformations on the test set. The reported accuracy is in percentage points.}
    \vspace{4pt}
    \begin{tabular}{p{7cm} r}
    	\toprule
    	\multicolumn{1}{c}{Method} & \multicolumn{1}{c}{Accuracy} \\
        \midrule
        Deformable with NCC, DWM, and MI \cite{Ferrante:16} & $69.23$ \\
        Proposed                                            & $\mathbf{84.62}$ \\
        \bottomrule
    \end{tabular}
    \label{tab:disease}
\end{table}

\section{Conclusion}
In this paper, we propose a novel method which exploits the 3D CNNs to calculate the optimal transformation (combining a linear and a deformable component within a coupled framework) between pair of images that is modular with respect to the similarity function, and the nature of transformation. The proposed method generates deformations with no self-crossings due to the way the deformation layer is defined, efficient due to the GPU implementation of the inference and reports high promising results compared to other unsupervised registration methods. Currently, the proposed network was tested on the challenging problem of lung registration, however, its evaluation on the registration of other modalities, and other organs is one of the potential directions of our method.

%\bibliographystyle{splncs04}
%\bibliography{book}

\end{document}